\documentclass[12pt,a4paper]{article}
\usepackage[utf8]{inputenc}
\usepackage[english]{babel}
\usepackage{amsmath}
\usepackage{amsfonts}
\usepackage{amssymb}
\usepackage{graphicx}
\usepackage{hyperref}
\usepackage{caption}
\usepackage{subcaption}
\usepackage{color}
\usepackage{enumerate}
\usepackage[margin=1.in]{geometry}
\usepackage{hyperref}
\title{Morpheo\\ Traceable Machine Learning on Hidden data}

\makeatletter
\renewcommand\@date{{%
  \vspace{-\baselineskip}%
  \large\centering
  \begin{tabular}{@{}c@{}}
    Mathieu Galtier\footnote{Rythm, Paris, France} \\
    \normalsize mathieu@rythm.co
  \end{tabular}%
  \quad \quad \quad
  \begin{tabular}{@{}c@{}}
    Camille Marini\footnote{CMAP, \'Ecole Polytechnique, Palaiseau, France} \\
    \normalsize camille.marini@polytechnique.edu
  \end{tabular}

  \bigskip
  \bigskip

  \today
}}
\makeatother

\begin{document}

\maketitle

\begin{abstract}
Morpheo is a transparent and secure machine learning platform collecting and analysing large datasets. It aims at building state-of-the art prediction models in various fields where data are sensitive. Indeed, it offers strong privacy of data and algorithm, by preventing anyone to read the data, apart from the owner and the chosen algorithms. Computations in Morpheo are orchestrated by a blockchain infrastructure, thus offering total traceability of operations. Morpheo aims at building an attractive economic ecosystem around data prediction by channelling crypto-money from prediction requests to useful data and algorithms providers. Morpheo is designed to handle multiple data sources in a transfer learning approach in order to mutualize knowledge acquired from large datasets for applications with smaller but similar datasets.
\end{abstract}

\noindent \textbf{Document purpose:}\\
This is a white-paper: an introduction to the technology powering \href{http://morpheo.co}{Morpheo}. It is evolving with the project to reflect the current vision and development status of Morpheo. It introduces Morpheo as a data-agnostic platform, although Morpheo is originally developed for a medical application, namely sleep data analysis. For technical details about Morpheo, see the \href{https://morpheoorg.github.io/morpheo}{online documentation and source code}.

\section{Context}

We are at a crucial moment in history where the progress in artificial intelligence is meeting an ever increasing computing power, and the collection of unprecedented amount of quality data. It is clear that this wave of digital intelligence is overwhelming many fields and in particular Healthcare. Not only is traditional medicine undergoing massive digitalization, but the development of medical-grade sensing products used everyday by the consumers is fuelling the emergence of predictive precision medicine. Thus, the perspective of mass medicine with a high degree of personalization becomes closer with these new technologies. Not only are we able to analyse and detect diseases at a large scale, but we will soon be able to identify slights variants we could not have observed to before. Sleep is particularly rich and under-explored source of information. Physiological signals such as brain and heart activity during sleep will lead to the definition of relevant biomarkers to characterize many pathologies from insomnia, sleep apnea to depression and neurodegenerative diseases. In few years, we will be able to detect pathologies before it even happens, to treat earlier, faster and cheaper. 

At the heart of this revolution lies Artificial Intelligence and, more precisely, \textbf{Machine Learning}. The central paradigm is to design models which learn by example: training algorithms on large datasets of examples builds models which can extrapolate to new data. With the appropriate algorithms \cite{lecun2015deep}, the quality of the models prediction is directly linked to the quality and quantity of labelled data used for training. Thus, most applications require large datasets to reach the state-of-the-art performance, as has been demonstrated in some sub-fields of medicine  \cite{esteva2017dermatologist}.

However, these huge quantities of collected \textbf{data are heterogeneous} and often need harmonization: they come from different sources thus with likely different qualities, resolutions, types of information. Some of this data may have been manually annotated, some may not. To cope with data heterogeneity, a new promising methodology is emerging: Transfer Learning \cite{Pan:2010, Yosinski:2014}. It allows merging datasets from different sources in a unique machine learning model which benefits from all data. In fact, several field are seeing the emergence of a unique deep neural network as a basis for other learning architectures, notably imagenet is the international standard for image processing \cite{krizhevsky2012imagenet}. To our knowledge, there is no such basis model in sleep data analysis.

A key topic for collecting so many data is \textbf{privacy}, even more with sensitive data as in Healthcare. There are severe ethical issues regarding the massive collection of private identifying data which is strictly regulated by requiring anonymization of such datasets \cite{g292014}. Beyond ethics, there are countless situations where having a macroscopic information is useful, yet the different actors do not necessarily want to make their data accessible to others. Concerning sleep data, pooling individual data would significantly increase the algorithms efficiency for various pathology detection, but there is a lot of reluctance of the actors to share openly their data, even more in a world with non-trusted actors which could reuse data for a different purpose. Thus, there is a strong need for a system putting sensitive data together, while guaranteeing their privacy and ownership.

Operations on sensitive data require a trusted environment which can be granted by a blockchain infrastructure. On traditional platforms, we rarely know how our data are going to be stored, processed and transmitted. The blockchain technology provides a radical solution to this major problem: total \textbf{transparency and traceability} of transactions. Beyond the simple exchange of crypto-currencies, a new generation of blockchain protocols, such as Ethereum \cite{Ethereum:2017} and Hyperledger \cite{Hyperledger:2016}, is putting forward the notion of self-enforcing smart contracts: if you subscribe to a smart-contract its execution is not avoidable and rigorously in the terms defined. With these systems, one does not need to trust the actors, because the protocol is preventing them from doing things differently from what is claimed.

Morpheo is a platform living at the intersection between machine learning, transfer learning, and blockchain. It intends to build a trustful environment for building traceable machine learning models while keeping the most advanced privacy possible for data providers. In the following, we will introduce the principles and the high level design of the Morpheo platform.

\section{Principles}

Morpheo is a for-privacy platform which attracts algorithms and data in order to provide accurate and automatic detection of patterns. It is intended to be used as a software as a service (SAAS) in business models based on the valorisation of state-of-the-art prediction. The code is open source.

The basic idea of Morpheo is to gather algorithms and data, while guaranteeing total privacy and ownership, in order to learn efficient machine learning models to detect a specific pattern in new data. For now, it exclusively performs supervised learning tasks where the patterns of interest have been manually identified in many examples.

Morpheo is data and algorithm agnostic (even beyond Machine Learning), but it is clear that the scope of the data has to be limited and clearly specified for each deployment of the platform. For instance, the first deployment of Morpheo is dedicated to sleep medicine based on sleep physiological recordings such as the electroencephalogram.

The efficiency of the approach is directly linked to the quantity and quality of data and algorithms handled by Morpheo. Therefore, it is crucial that Morpheo offers an attractive environment such that data owner and algorithm developers naturally tend to provide their data to the Morpheo platform. As detailed below, Morpheo implements strong privacy features and fair economic retribution to enforce this attractivity.

\subsection{Data-centric Machine Learning}

Morpheo is basically a Machine Learning backend which automates the learning and prediction of various algorithms on multiple data sets. The platform handles the algorithms source code and apply it to various hidden data remotely. At the core of the approach is the systematic update of a benchmark table, to identify which algorithm performs better on a specific prediction challenge. Even if an algorithm does not use Machine Learning, it can be compared to others in a transparent way.

Morpheo's design philosophy is data-centric: we believe it is better to invest effort on a few algorithms with lots of data, as opposed to many algorithms with a moderate amount of data. This partially mitigates the quadratic explosion of computations when there are both many algorithms and much data. Thus, Morpheo will deliberately favor the few best algorithms by feeding them with new data in priority. In the end, there may be only a single algorithm performing much better than the aggregate of the others.

Eventually, Morpheo will have the ability to handle seamlessly different sources of data for a single problem: it is designed as a tool for Transfer Learning. To aggregate as much data as possible, Morpheo's infrastructure is adapted to manipulating various data formats, various distributions and potentially with missing data. It is up to the algorithm developers to leverage this flexibility with adapted Transfer Learning approaches.

\subsection{For-privacy principles}

Morpheo provides the highest degree of privacy for data and algorithm source code. Morpheo is designed specifically to handle sensitive data which must remain hidden from both the algorithm developers, but also the parties which participate to the deployment of the infrastructure. Morpheo implements the following high-level principles:
\begin{enumerate}
\item \textbf{Respect of ownership} of data and algorithms. No human or unwanted algorithm can ever access the data and/or algorithms of others.
\item \textbf{Transparency and traceability} of data and algorithm use. All the operations done securely on the data and algorithms, with the agreement of the owner, are logged in a public ledger. The self-enforcing mechanisms orchestrating the operations are public.
\end{enumerate}

The trade-off between privacy and transparency will be done by hiding systematically the content of the data, but allowing inspections of the full list of operations made on the data without revealing their result.

\subsection{Economic incentive}

In order to attract as many useful data and algorithms as possible, Morpheo relies on a transparent retribution system for information providers.

Being backed by a blockchain, Morpheo intrinsically relies on a specific crypto-currency to quantify the economic value of automated predictions. The convertibility of such crypto-money with real money is neither relevant nor pursued at this stage of the project. A strong benefit of the blockchain is that transactions are entirely transparent: anybody can consult the self-enforcing rules governing the crypto-money flow.

Morpheo is built as an economic circuit. Users can 'pay' for requesting predictions by providing a certain amount of crypto-money which is then fairly split between useful data and algorithms providers. It creates an incentive to provide data since it mechanically gives access to more predictions requests. As in most blockchain infrastructures, those contributing to the system infrastructure and security will also get fees on the processed transactions.

Because more data leads to more crypto-money, we need to make sure bad data are not blindly added to the platform. Morpheo transparently computes a global contributivity score for each data and algorithm which will be the basis for retribution. Computing the contributivity of a single datum amounts to comparing the performance of various models having been learned with and without this datum. If adding the datum leads to a sharp increase in performance, then it will have a high contributivity. Useless data will have zero contributivity and will not be used for subsequent model training leading to no retribution.

\subsection{Use case}

Morpheo users want to get accurate predictions on new data. Designed with Healthcare application in mind, Morpheo helps medical experts and doctors analyse raw data and provide suggestions for diagnosis. The applications to other fields dealing with sensitive data appear numerous. Morpheo can also be used as a backend for external web services, IoT devices or connected medical devices to provide immediate predictions or pattern detection.

In the current development phase, we focus on a typical expert who wants to accelerate her data annotation process. She would upload her data and ask for an automated prediction. Then by visualizing the data, she would check the quality and, sometimes, correct the annotation. Thus, we can provide a time-saving service and improve our algorithm quality at the same time. This use case is particularly appropriate for the field of sleep medicine where night analysis is particularly time consuming.

\section{Design}

Morpheo is an open source initiative to build a complex suite of services from a modern front-end, and a heavy computation backend to a blockchain infrastructure. Starting from user interactions, we now describe the architecture and the main ideas for encryption.

\subsection{User interactions}

The user interface will be a cross-platform software made freely available to anyone. It will necessitate an internet connection for data transfer and security protocols.

\noindent The main actions that the user will be doing are:
\begin{itemize}
\item Importing and exporting data. It will be made as simple as possible, for instance with a simple drag and drop procedure and clear permission features. Morpheo will provide Compatibility with common data formats.
\item Visualizing the data and intermediary metrics through a user-friendly interface. For sleep, Morpheo represents the filtered signals (EEG, EOG, EMG, ECG) on epoch of 30 seconds and advanced metrics on sleep stages.
\item Annotating data. Morpheo will integrate an intuitive way to provide this information to the platform, for instance, with annotation tools integrated in the visualization of the raw signals to identify specific events.
\item Inputting algorithms. A command line interface and a feature rich text field will be provided to input algorithms. A detailed benchmark of all algorithms will be accessible to compare their efficiency. Data samples will be provided for prototyping.
\item Requesting predictions. The requested prediction will be displayed appropriately within the visualization interface, so that the user can correct the algorithm when it has made a mistake.
\end{itemize}

\subsection{Platform architecture}

\begin{figure}
    \centering
    \begin{subfigure}[b]{0.49\textwidth}
        \includegraphics[width=\textwidth]{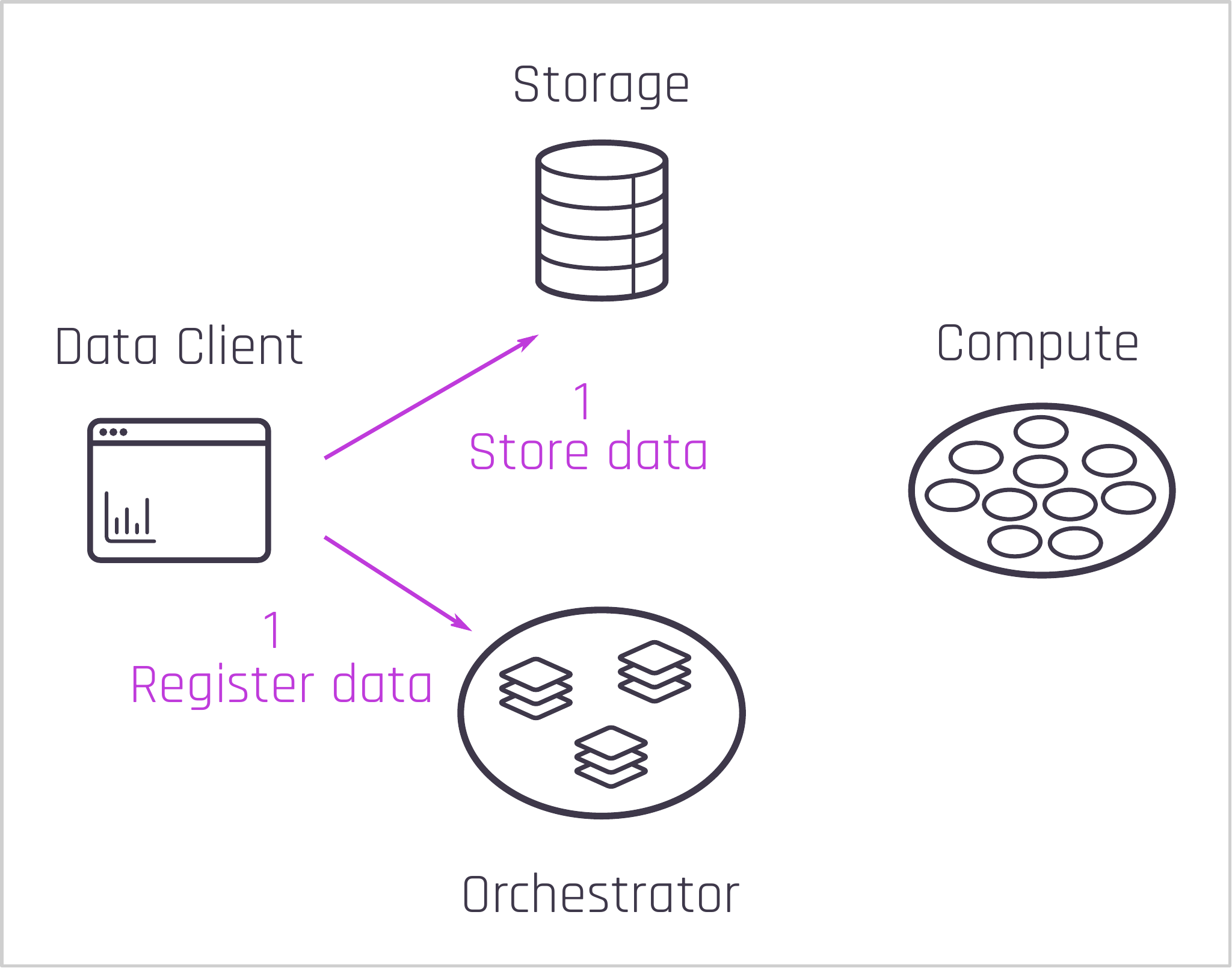}
        \caption{\textbf{Step 1}: A user uploads a new \textit{data}. It is encrypted on the \textit{data client} and sent to \textit{Storage}. \textit{Data} are registered on \textit{Ochestrator} which adds several tasks to the ledgers.}
        \label{fig:step1}
    \end{subfigure}
    \begin{subfigure}[b]{0.49\textwidth}
        \includegraphics[width=\textwidth]{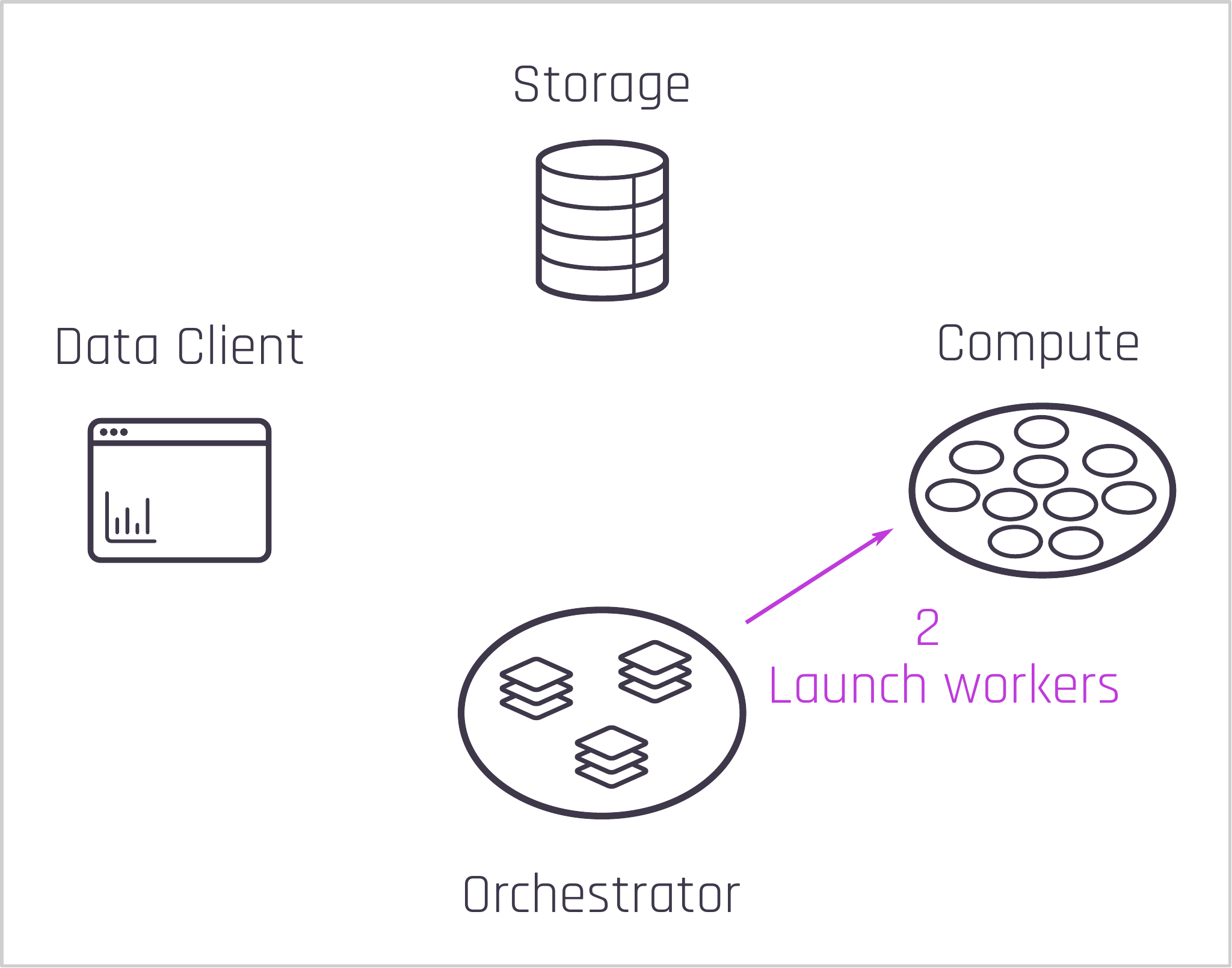}
        \caption{\textbf{Step 2}: A worker is launched to consume the task. The worker id (a public key) is written in the ledgers. The associated private is hosted on the worker to prove its identity.}
        \label{fig:step2}
    \end{subfigure}
    \begin{subfigure}[b]{0.49\textwidth}
        \includegraphics[width=\textwidth]{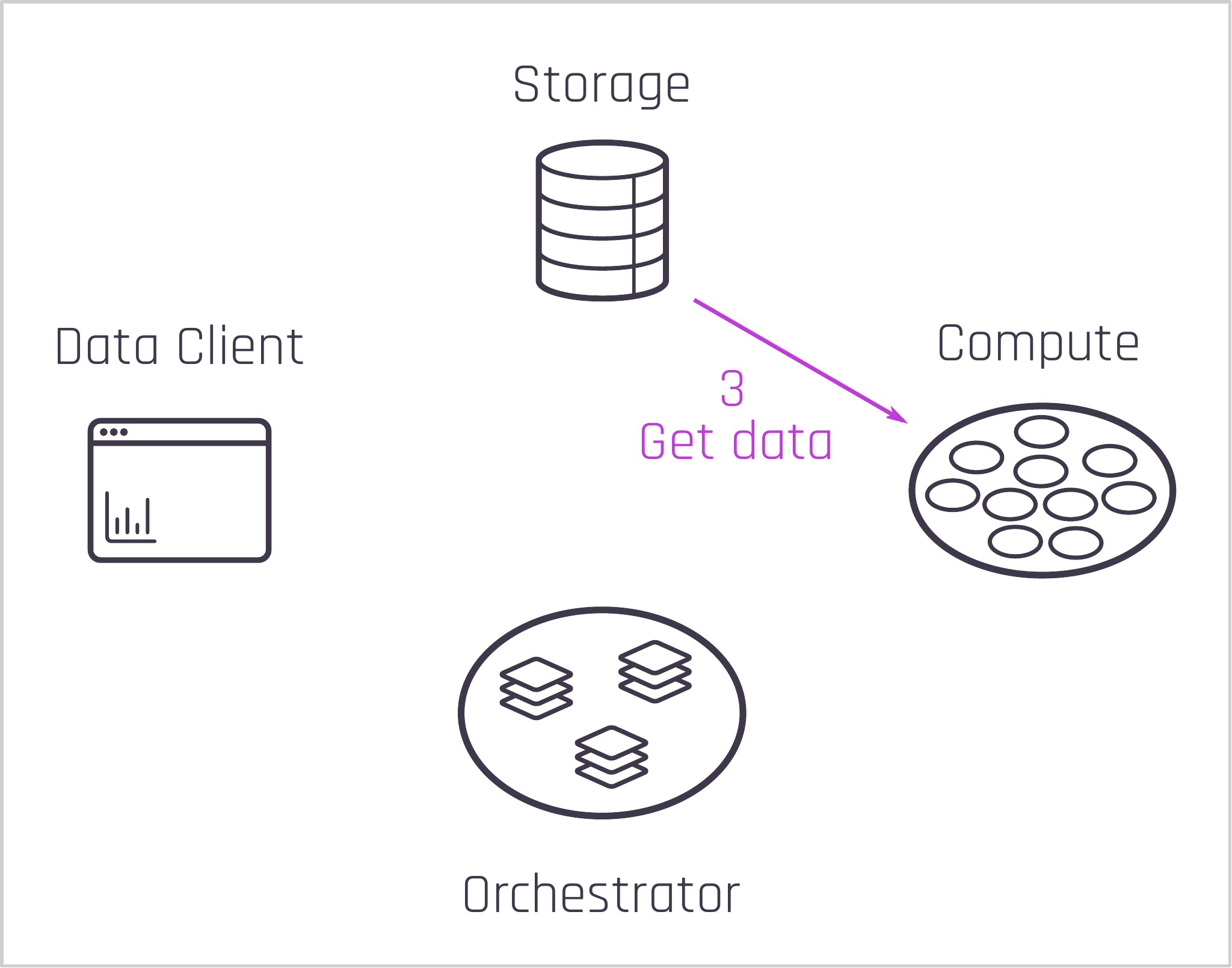}
        \caption{\textbf{Step 3}: The worker gets encrypted \textit{data} and their decryption keys. Workers are authenticated via their id in the \textit{Orchestrator} ledger.}
        \label{fig:step3}
    \end{subfigure}
    \begin{subfigure}[b]{0.49\textwidth}
        \includegraphics[width=\textwidth]{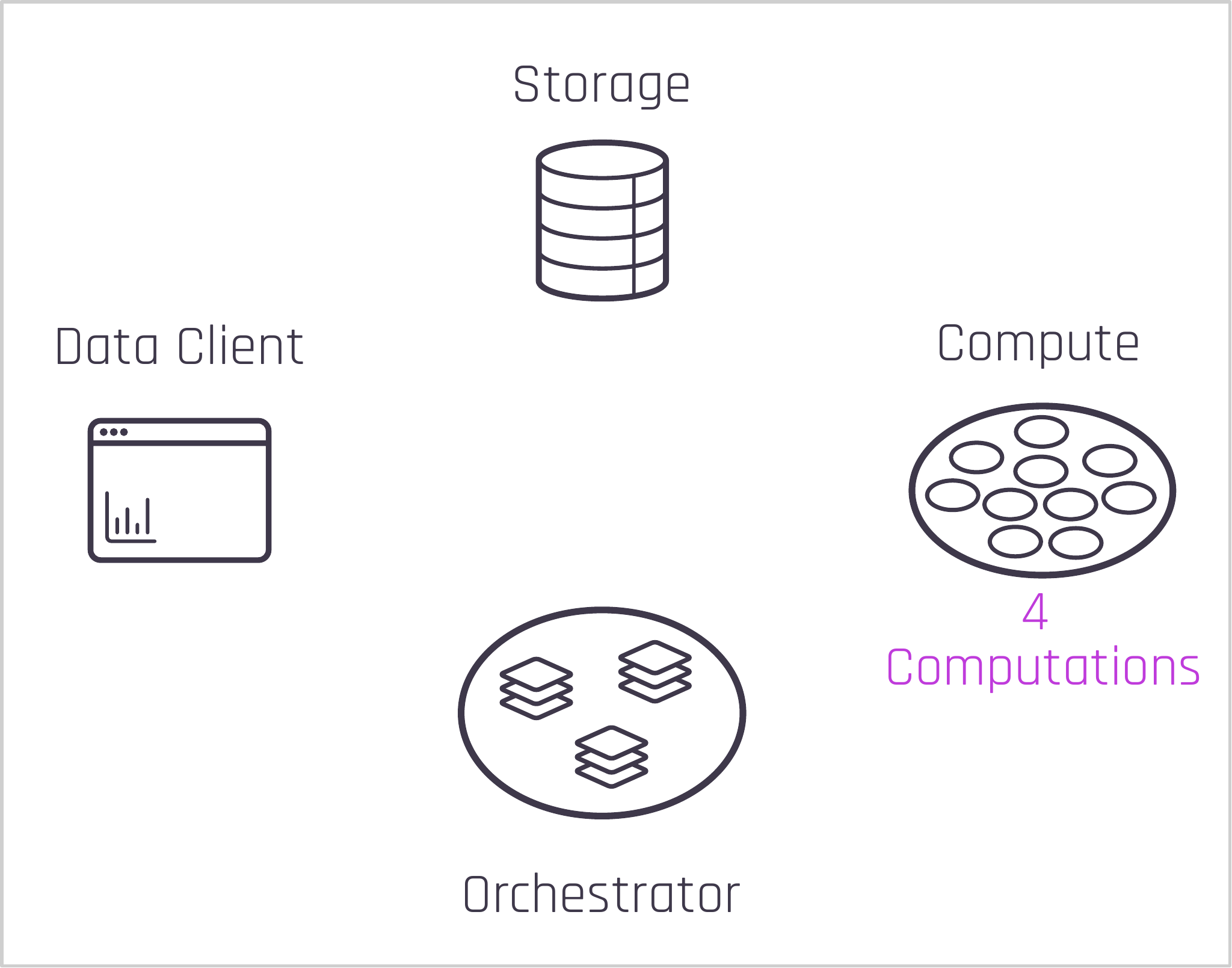}
        \caption{\textbf{Step 4}: The worker decrypts the data and does learning / computes predictions with no connection to the rest of the world.}
        \label{fig:step4}
    \end{subfigure}
        \begin{subfigure}[b]{0.49\textwidth}
        \includegraphics[width=\textwidth]{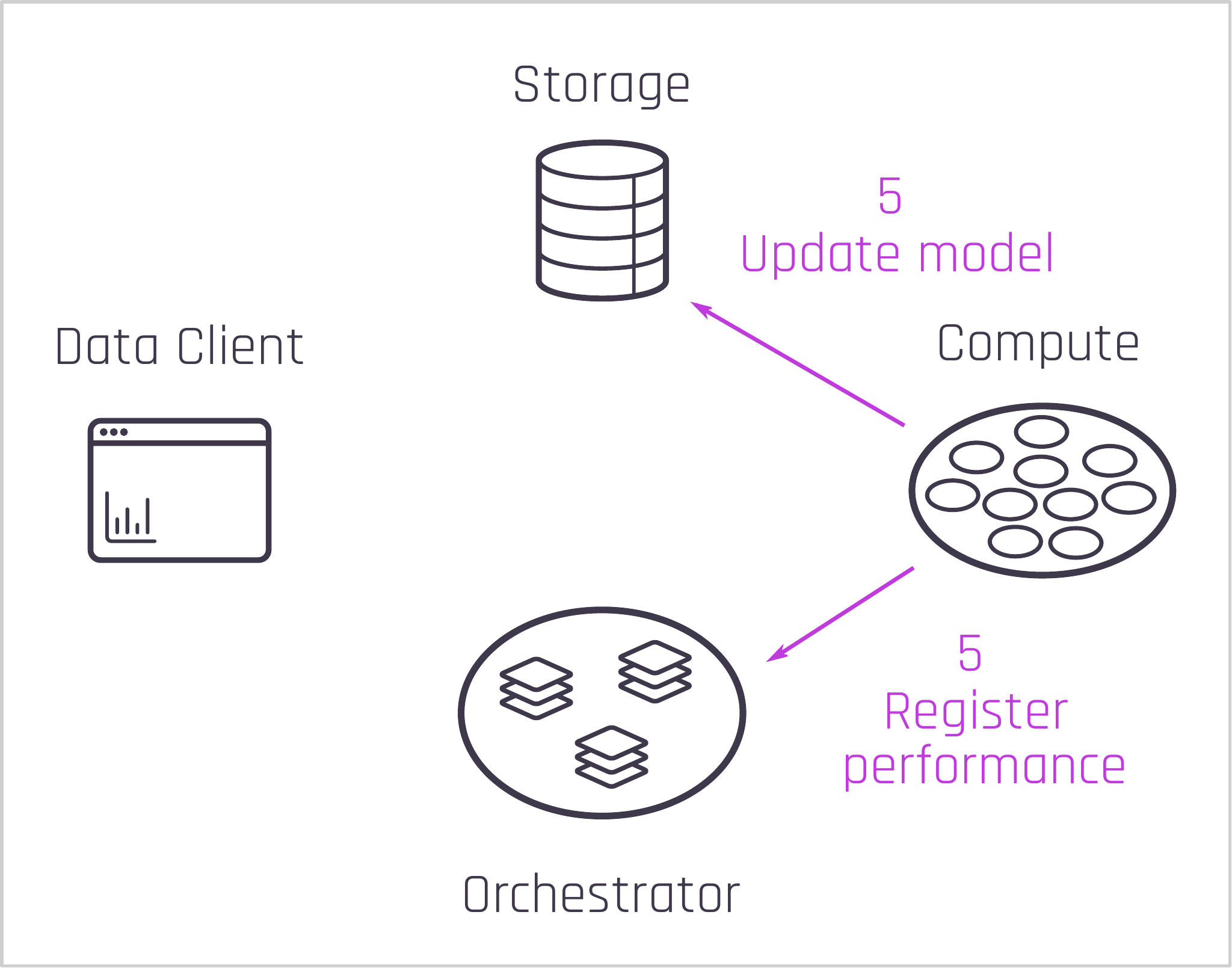}
        \caption{\textbf{Step 5 (learn)}: The worker returns the performance and updates the model.}
        \label{fig:step5learn}
    \end{subfigure}
    \begin{subfigure}[b]{0.49\textwidth}
        \includegraphics[width=\textwidth]{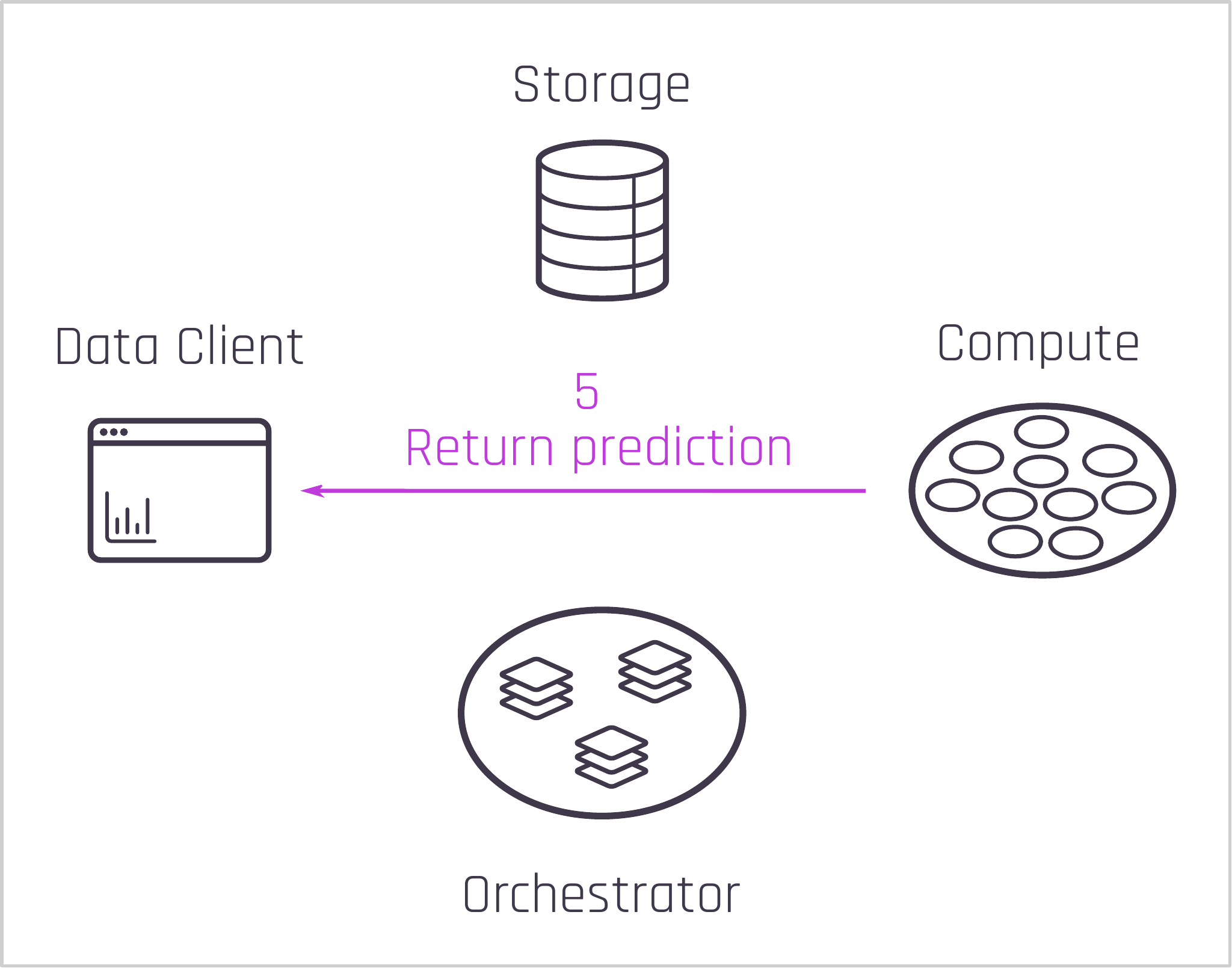}
        \caption{\textbf{Step 5 (predict)}: The worker returns the prediction to the user without revealing it.}
        \label{fig:step5predict}
    \end{subfigure}
    \caption{Workflow of a learning or prediction task. Step 1 to 4 are identical for both cases.}\label{fig:architecture}
\end{figure}

To guarantee data and algorithms privacy and retribution, while combining them to create state-of-the-art prediction algorithms, we have designed an advanced infrastructure for Morpheo. It is made of 4 different parts: (i) a local client running on data providers machines, (ii) a cloud storage, (iii) cloud computation resources and, (iv) a private blockchain network. The functioning of this architecture is illustrated in figure \ref{fig:architecture}.

In this section and for simplicity, we will refer to both raw data, algorithms source code, and machine learning models (post-learning) simply as \textit{data}. Indeed, their processing is very similar and we will assume it is identical except when explicitly mentioned otherwise.

A core design philosophy of Morpheo's architecture is that the \textit{data} should remain encrypted as much as possible, in particular the \textit{data} are always stored encrypted. Decryption keys are stored locally on the \textit{data} owner's computer. With this simple infrastructure, a user can already securely store and visualize her \textit{data}. This motivates the definition of two components of the platform:
\begin{enumerate}[(i)]
\item \textit{Data client} is a fat-client installed on users computers, which can en/decrypt \textit{data}, upload/download encrypted \textit{data} to \textit{Storage}, visualize \textit{data}, and store decryption keys.
\item \textit{Storage} is a simple database where \textit{data} are always encrypted. Although there would be no fundamental problem in distributing it, there are also no risk centralizing it since the host does not have access to the decrypted \textit{data}.
\end{enumerate}

Yet, to perform Machine Learning tasks, such as prediction and learning, the \textit{data} must be decrypted somehow. The chosen solution is to create secure disposable virtual computers for each task, which we call workers. The encrypted \textit{data} are downloaded on the worker, which then requests the keys to decrypt the \textit{data} locally. Then it performs the given task and returns the results. Finally, it self destructs, erasing all \textit{data} and decryption keys. This corresponds to the following component of the architecture:
\begin{enumerate}[(i)]
\setcounter{enumi}{2}
\item \textit{Compute} is creating, managing and certifying a swarm of workers which are ephemeral sorts of virtual machines which aim at performing learning or prediction tasks. Most of the computing time of the architecture is concentrated in this component. For simplicity, we use centralized resources for now.
\end{enumerate} 

To complete the architecture, it is necessary to trustfully control who can access the \textit{data} and for what purpose. We need an infrastructure-wide certifying authority which associates \textit{data} and workers, allowing the latter to decrypt the former. This is the trust bottleneck which cannot be compromised. In a way, all the architecture is built around this component which is the traceability, and security cornerstone:
\begin{enumerate}[(i)]
\setcounter{enumi}{3}
\item \textit{Orchestrator} is transparently organizing the computations performed by the platform by combining valuable \textit{data} to get the best performance possible. Most importantly, it creates and maintains a ledger of all learning operations in the (simplified) form:
\begin{equation*}
\left[\mbox{raw data id}, \mbox{model id}, \mbox{worker id}, \mbox{performance}\right]
\end{equation*}
where the 'id' are identification strings. Chronologically, it first creates such a quadruplet with an empty performance, serving as a job queue for \textit{Compute} (registration of \textit{data} as in step 1 of figure \ref{fig:architecture}). Second, when the task has been performed by \textit{Compute}, a performance value is returned and added to the learning quadruplet (registration of performance as in 'step 5 learn' of figure \ref{fig:architecture}). A similar ledger (without performance) is stored for prediction tasks. Third, it updates the contributivity of \textit{data} directly from the learning ledger. \textit{Orchestrator} runs on a private blockchain with smart contracts providing a trustful proof that the \textit{data} have only been used for there agreed upon purpose.
\end{enumerate}

\subsection{\textit{Data} encryption}

\textit{Data} are always encrypted symmetrically on the \textit{data} client before being transmitted to the rest of the platform. The symmetric key is always stored securely on the client side and will make it possible to decrypt data locally for visualization and annotation.

The decryption keys are then shared on the private network running the blockchain. Indeed, it cannot be assumed that the \textit{data client} can always provide the \textit{data} decryption keys to the other components of the platform (e.g. the \textit{Compute} workers). This would imply that the clients have to be always online to provide keys. To circumvent such a strong constraint, the fat-client can use a multi-party encryption scheme to spread its keys on the private network nodes so that the \textit{data} can be decrypted if all the nodes provide partial keys to \textit{Compute}. As a simple implementation, the symmetric decryption key could be split in several parts which are sent to different nodes in the network. In this improved version, the \textit{data} owner can unplug its computer without disrupting the platform, while being sure that the \textit{data} cannot be retrieved by a single node. This means that the network private nodes not only run a blockchain protocol but also provide a service for securely receiving and sending the key parts. The \textit{Orchestrator} plays a central role in the authentication of the workers to guarantee to the nodes that they are trustful: nodes only send their key part to a worker which can prove with its private key that its public key is written in one of the \textit{Orchestrator} ledgers.

In Machine Learning, raw data and algorithms are combined into models that perform predictions. In Morpheo, it is considered that these models do not belong to anyone since they derive from various \textit{data} which are themselves retributed through the contributivity mechanism. The models themselves are simply seen as by-products of the platform, which remain encrypted and inaccessible to anyone, except the platform itself as described in the previous paragraph. In other words, anybody can request a prediction through the platform and use the models, but nobody can access them outside this framework.

\subsection{Release agenda}

Morpheo is under heavy development and currently unstable. Open sourcing of all Morpheo Component is expected in Spring 2017. As a first prototype release in Fall 2017, we focus on a functional platform with a classical backend for the \textit{Orchestrator}. The first stable release with a blockchain backend is expected in Spring 2018.

\section{Discussion}

We have introduced Morpheo, a transparent machine learning backend with a strong privacy for data and algorithms. Its main goal is to provide automated and accurate pattern predictions. It systematically trains the algorithms and provides transparent benchmarks of models performances. It is built to perform transfer learning so as to make the most of various sources of data it collects. Morpheo is open source and promotes new, open, decentralized organisations of computation. It aims at being entirely transparent and secure so as to provide a trusted and attractive environment for data providers.

Morpheo hosts a prediction market where useful data and algorithms are rewarded. By computing the contributivity of each datum, Morpheo is able to divide fairly the price paid by a user to request a prediction on his data. The platform uses its own crypto-currency. Eventually, Morpheo aims at creating an economically attractive environment for data providers and prediction requesters. This will guarantee the prediction accuracy of the Morpheo approach.

Although Morpheo is built to transparently protect data privacy, there are several risks associated to the approach. To prevent attackers from uploading bad data, we compute a contributivity score which cancels the interest in upload anything but useful data. To mitigate the risk of a malicious algorithm stealing data from \textit{Compute}, we are implementing advanced security measures and notably the inability of the worker to communicate when the algorithm is being run. The risk corresponding to a malicious host is considered seriously and limited by the private nature of the network and a demanding certification procedure imposed by the consortium. Finally, we may consider a distributed decryption key backup, to recover data when the keys have been lost.

The fields of application of Morpheo are numerous: situations where data are so sensitive that actors are not sharing them yet are potential interesting fields for Morpheo. Indeed, we provide a secure way to extract information and predictors from the statistical information without revealing the individual data. In Healthcare, data are still kept private in large data banks and Morpheo could be an economic opportunity for such actors to valorize their assets. Medically, Morpheo could help diagnose several pathologies from skin diseases, to cancer detection and anticipation. Beyond Healthcare, Morpheo could be applied to fields where the economic system is set such that actors do not wan to share their information. Not only does it apply remarkably to finance, but also more generally when there is a need for a market-wise automated decision making based on competing actors.

\section*{Acknowledgements}

Morpheo/Dreemcare is funded by the French state through a strategic investment programme (Programme d'Investissement d'Avenir) and by Rythm.

\bibliographystyle{unsrt}
\bibliography{biblio.bib}
\end{document}